\documentclass{article}

\PassOptionsToPackage{numbers, compress}{natbib}

\usepackage[preprint]{neurips_2024}

\usepackage[utf8]{inputenc} 
\usepackage[T1]{fontenc}    
\usepackage{url}            
\usepackage{booktabs}       
\usepackage{amsfonts}       
\usepackage{nicefrac}       
\usepackage{microtype}      
\usepackage[table,xcdraw]{xcolor}         

\usepackage{bm}
\usepackage{wrapfig}
\usepackage{caption}
\usepackage{graphicx}
\usepackage{multirow}
\usepackage{amsmath}
\usepackage{enumitem}
\usepackage{listings}
\usepackage{algorithm}
\usepackage{algorithmic}
\usepackage{longtable,array}
\newcolumntype{M}[1]{>{\centering\arraybackslash}m{#1}}

\usepackage[pagebackref=true,breaklinks=true,letterpaper=true,colorlinks,bookmarks=false]{hyperref}

\title{Peer Review as A Multi-Turn and Long-Context Dialogue with Role-Based Interactions}

\author{%
Cheng Tan$^{1*}$, Dongxin Lyu$^{2}$\thanks{Equal contribution.}, Siyuan Li$^{1*}$, Zhangyang Gao$^{1}$, \\ \textbf{Jingxuan Wei$^{3}$, Siqi Ma$^{1}$, Zicheng Liu$^{1}$ and Stan Z. Li}$^{1}$\thanks{Corresponding author.} \\
$^{1}$Westlake University, $^{2}$Jilin University, $^{3}$University of Chinese Academy of Sciences \\
 \\
}

\begin{document}

\maketitle
\begin{abstract}
Large Language Models (LLMs) have demonstrated wide-ranging applications across various fields and have shown significant potential in the academic peer-review process. However, existing applications are primarily limited to static review generation based on submitted papers, which fail to capture the dynamic and iterative nature of real-world peer reviews. In this paper, we reformulate the peer-review process as a multi-turn, long-context dialogue, incorporating distinct roles for authors, reviewers, and decision makers. We construct a comprehensive dataset containing over 26,841 papers with 92,017 reviews collected from multiple sources, including the top-tier conference and prestigious journal. This dataset is meticulously designed to facilitate the applications of LLMs for multi-turn dialogues, effectively simulating the complete peer-review process. Furthermore, we propose a series of metrics to evaluate the performance of LLMs for each role under this reformulated peer-review setting, ensuring fair and comprehensive evaluations. We believe this work provides a promising perspective on enhancing the LLM-driven peer-review process by incorporating dynamic, role-based interactions. It aligns closely with the iterative and interactive nature of real-world academic peer review, offering a robust foundation for future research and development in this area. We open-source the dataset at \href{https://github.com/chengtan9907/ReviewMT}{github.com/chengtan9907/ReviewMT}.
\end{abstract}

\section{Introduction}

Language models (LMs) serve as a cornerstone in the field of artificial intelligence and its applications~\cite{gao2022pifold,li2023moganet,tan2022simvp}. The introduction of the highly parallelizable Transformer model ~\cite{vaswani2017attention} marked a significant milestone. This breakthrough was exemplified by the development of BERT~\cite{devlin2018bert}, which revolutionized the field by introducing pre-training bidirectional language models on large-scale unlabeled corpora using specially designed tasks. BERT's success established the pre-training and fine-tuning paradigm, inspiring a wave of subsequent research and advancements in pre-trained language models (PLMs)\cite{radford2019language,liu2019roberta,lan2019albert,sanh2021multitask,fedus2022switch,wang2022language}. Scaling up these pre-trained models led to the development of large language models (LLMs)~\cite{kaplan2020scaling,wei2022emergent,zhou2023comprehensive,han2021pre,liu2023pre}. This progression culminated in the creation of ChatGPT and GPT-4\cite{gpt4}, which have demonstrated unprecedented performance across language tasks. The technical evolution of LLMs has garnered increasing popularity in both industry and academia, with applications spanning a wide range of fields, including healthcare~\cite{tang2023does,yang2023evaluations,singhal2023towards,yang2024zhongjing,chen2023utility}, finance~\cite{wang2023fingpt,shah2023zero,xie2024finben,zhao2024revolutionizing}, education~\cite{szabo2023chatgpt,malinka2023educational,susnjak2022chatgpt}, and scientific research~\cite{jin2019pubmedqa,tan2021co,fang2023mol,wei2023enhancing,tan2023boosting,amin2023will,tan2024retrieval,park2024can}.

\begin{figure}[h]
  \centering
  \includegraphics[width=1.0\textwidth]{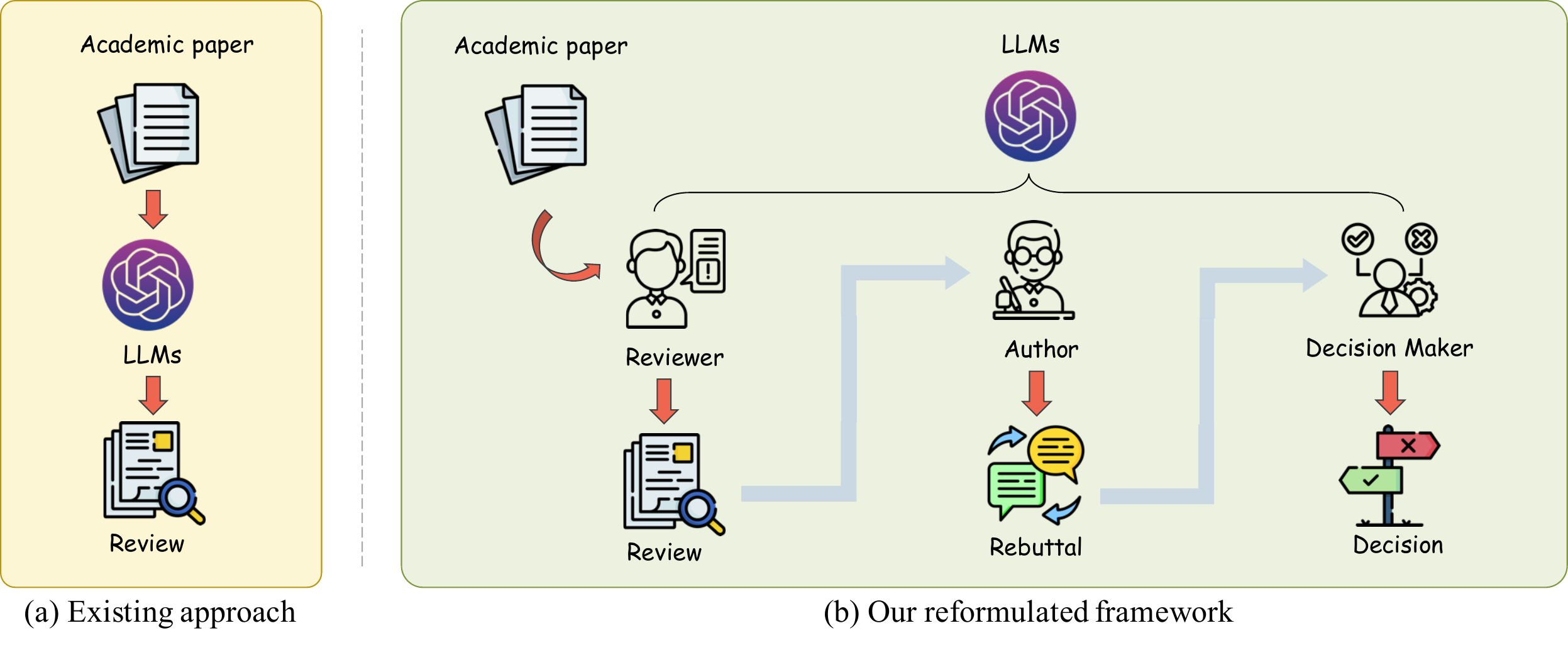}
  \caption{Comparison of existing LLM applications in peer review and our reformulated framework.}
  \label{fig:motivation_comparison}
\end{figure}  

Academic paper peer-review is a critical component of the academic publishing system, ensuring the quality of scientific research. Despite its essential role, the traditional peer-review process faces significant criticism~\cite{morris2023using,shah2022challenges,liu2023reviewergpt} for its inefficiency, bias, and lack of transparency. While applying LLMs in peer-review presents a promising solution, recent studies have demonstrated the potential of LLMs in generating high-quality reviews for given papers~\cite{robertson2023gpt4,liang2023large,darcy2024marg}. However, existing research primarily focuses on generating static reviews based on submitted papers, which severely simplifies the process and fails to capture the dynamic and iterative nature of real-world peer reviews. In this work, as shown in Figure~\ref{fig:motivation_comparison}, we offer a novel perspective on the complete peer-review process by reformulating it as a multi-turn, long-context dialogue involving three distinct roles: authors, reviewers, and decision makers. This reformulation includes several key aspects:
\begin{itemize}[leftmargin=5mm]
  \item \textbf{Long-Context}: The entire dialogue is grounded in the extensive context of the paper, ensuring that all interactions of different roles are informed by the full scope of the manuscript.
  \item \textbf{Multi-Turn}: The dialogue is conducted over multiple rounds, mimicking the real world where reviewers write reviews based on the paper, authors provide rebuttals to the reviews, reviewers respond to rebuttals, and decision makers make decisions based on the comprehensive exchange.
	\item \textbf{Role-Based}: Each role in the dialogue has specific responsibilities and objectives. Reviewers critically evaluate the paper and provide feedback, authors respond to this feedback to clarify their work, and decision makers synthesize the dialogue to make an informed publication decision.
\end{itemize}
With these principles in mind, we constructed a comprehensive dataset named \texttt{ReviewMT}, sourced from multiple venues including the top-tier AI conference ICLR and the multidisciplinary journal Nature Communications. This dataset is meticulously designed to embody the dynamic, iterative nature of the peer-review process. By incorporating both accepted and rejected papers from ICLR, the dataset provides insights into common pitfalls and areas for improvement, enriching the training and evaluation of LLMs. The dataset spans a wide range of domains, reflecting the diverse topics covered by Nature Communications and the cutting-edge AI research presented at ICLR. Each entry in the collected dataset is carefully annotated to include multi-turn dialogues that capture the full scope of interactions between authors, reviewers, and decision makers.

Creating the dataset is just the first step in our reformulated peer-review framework. To evaluate LLM performance in this setting, we propose a series of metrics tailored to each role in the dialogue. These metrics assess the validity of generated responses, the text quality, the score evaluation of final reviews, the decision evaluation of decision makers. By evaluating LLMs based on these metrics, we aim to provide a fair and comprehensive assessment of their performance in the peer-review process. We believe this work offers a promising perspective on enhancing the LLM-driven peer-review process by incorporating dynamic, role-based interactions. It closely aligns with the iterative and interactive nature of real-world academic peer review, providing a foundation for future research.

\section{Related Work}

\subsection{Instruction Tuning Dataset}

Instruction tuning is a specialized training process applied to LLMs to enhance their ability to follow specific instructions and perform designated tasks with greater precision and reliability. Instruction tuning datasets, which are collections of task-specific examples paired with explicit instructions, play a crucial role in this process. Early efforts in this field, such as Dolly~\cite{conover2023free} and InstructGPT~\cite{ouyang2022training}, relied heavily on manual or expert annotations. Over time, the field has seen the emergence of semi-automated and fully automated approaches for instruction creation. These approaches have transformed existing datasets and facilitated more efficient training of LLMs~\cite{chowdhery2023palm,sanh2021multitask,chung2024scaling}. A notable example is Stanford Alpaca~\cite{taori2023stanford}, which employs a bootstrapping technique grounded in a set of handcrafted instructions to generate 52,000 diverse instructions. This approach has inspired the development of model-aided data collections, such as Baize~\cite{xu2023baize}, COIG~\cite{zhang2023chinese}, and UltraChat~\cite{ding2023enhancing}, enabling automatic data generation and reducing the need for human effort~\cite{honovich2022unnatural,nayak2024learning,rajpurkar2016squad,wang2018glue}. Though these datasets have significantly advanced the instruction tuning field, most of them focus on single-turn interactions and lack the multi-turn, long-context dialogue characteristic of peer reviews.

\subsection{LLM in Review}

LLMs have demonstrated significant potential in reviewing and comprehending complex articles. Early studies~\citep{robertson2023gpt4} suggested that GPT-generated reviews are comparable to those of human reviewers. By comparing reviews generated by humans and GPT models for academic papers submitted to a major machine learning conference, it was initially demonstrated that LLMs can effectively contribute to the peer review process. Further research~\cite{liang2023large} revealed that GPT-4's feedback had a substantial overlap with human reviewers, with over half of the users rating GPT-4's feedback as helpful, underscoring the growing role of LLMs in the peer review process. MARG~\cite{darcy2024marg} employs multiple LLM instances to internally discuss and assign sections of a paper to different agents, providing comprehensive feedback across the entire text, even for papers exceeding the model's context size. While current LLM research focuses on simply generating reviews, we aim to simulate the complete review process into a multi-round dialogue, emphasizing the iterative nature of real-world peer review.

\section{Preliminaries}

In existing works on LLM-based peer review research, the focus is primarily on generating a static review $R$ for a given paper $P$ by a reviewer $\mathcal{R}$. This process can be formulated as follows:
\begin{equation}
  \mathcal{R}: P \rightarrow R,
\end{equation}
where $\mathcal{R}$ is implemented by an LLM $\mathcal{F}_\theta$ parameterized by $\theta$. This process is typically conducted in a single turn, with the reviewer $\mathcal{R}$ providing feedback on the paper $P$ without further interaction. 

In our peer-review framework, we extend this process to a multi-turn dialogue with three distinct roles: Reviewer, Author, and Decision Maker. Each role has specific objectives and interactions:
\begin{itemize}[leftmargin=5mm]
  \item \textbf{Reviewer} ($\mathcal{R}$): The reviewer is responsible for generating an initial review $R_i$ for the paper $P$ in the first turn, which includes a critical assessment of the paper and questions for the author to address: $\mathcal{R}_i: P \rightarrow R$. \textit{It is worth noting that there are $N$ reviewers for each paper $P$.}
  After the author responds with rebuttals $A_i$ in the second turn, the reviewer evaluates the author's response and generates a final review $R_i'$, which reflects their updated opinion on the paper after considering the author's clarifications and revisions: $\mathcal{R}_i: A_i \rightarrow R_i'$. 
  \item \textbf{Author} ($\mathcal{A}$): The author plays a crucial role in the second turn by responding to the initial review $\{R_i\}_{i=1}^N$ provided by each reviewer. The author carefully addresses the reviewer's comments, clarifies misunderstandings, and outlines any changes or improvements made to the paper in response to the feedback. The rebuttal $A_i$ serves to defend the paper's validity and significance while showing a willingness to incorporate constructive criticism: $\mathcal{A}: \{R_i\}_{i=1}^N \rightarrow \{A_i\}_{i=1}^N$.
	\item \textbf{Decision Maker} ($\mathcal{D}$): The decision maker synthesizes the entire dialogue, including the paper $P$, the initial review $\{R_i\}_{i=1}^N$, the author's rebuttal $\{A_i\}_{i=1}^N$, and the final review $\{R_i'\}_{i=1}^N$, to make an informed decision $D$. This role is pivotal in the fourth turn, where the decision maker evaluates the coherence and validity of the arguments presented by both the reviewer and the author to reach a final decision on whether the paper should be accepted or rejected: $\mathcal{D}: \{R_i, A_i, R_i'\} \rightarrow D$.  
\end{itemize}
The complete process can be formulated as a multi-turn dialogue with the following interactions:
\begin{equation}
\begin{aligned}
  \textcircled{1}\;&\{\mathcal{R}_i(P)\} \rightarrow \{R_i\} \\
  \textcircled{2}\;&\{\mathcal{A}(R_i)\} \rightarrow \{A_i\} \\
  \textcircled{3}\;&\{\mathcal{R}_i(A_i)\}\rightarrow \{R_i'\} \\
  \textcircled{4}\;&\mathcal{D}(\{R_i, A_i, R_i'\})\rightarrow D.
\end{aligned}
\end{equation}
By incorporating these roles into a multi-turn dialogue, our framework stimulates the dynamic and iterative nature of real-world peer review. This approach facilitates more detailed and interactive reviews, encouraging constructive communication between authors and reviewers.

\section{\texttt{ReviewMT} Dataset}

\subsection{Data Source}

One of the primary sources for the \texttt{ReviewMT} dataset was the International Conference on Learning Representations (ICLR)~\cite{iclr} on OpenReview~\cite{soergel2013open}, renowned for its contributions to machine learning and artificial intelligence. ICLR's emphasis on cutting-edge research across a broad spectrum of AI topics provided a diverse and valuable set of papers and reviews. Additionally, data was collected from Nature Communications~\cite{nature_comms}, a leading multidisciplinary journal known for publishing significant scientific advances. The broad scope and high impact of this journal allowed us to incorporate papers from various scientific fields, enhancing the dataset's diversity. Both ICLR and Nature Communications offer accessible and detailed review data, making them ideal sources for constructing a comprehensive peer-review dataset. Given the distinct review processes of these sources, the dataset is divided into two subsets: \texttt{ReviewMT-ICLR} and \texttt{ReviewMT-NC}. This division allows for tailored analysis and training, reflecting the unique characteristics and standards of each review process.

\subsection{Data Processing}

\begin{figure}[h]
  \centering
  \includegraphics[width=1.0\textwidth]{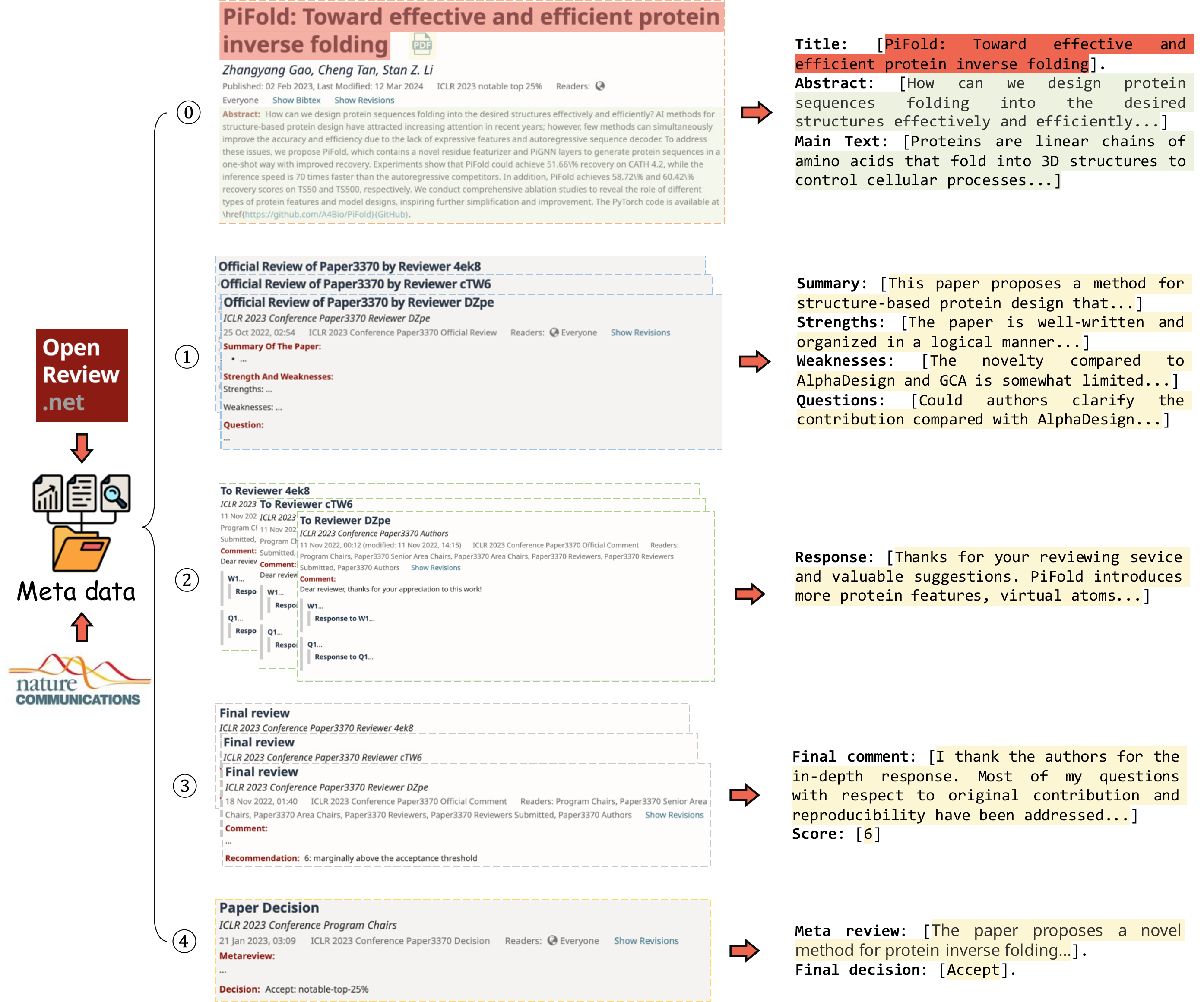}
  \caption{Overview of the data processing pipeline for the \texttt{ReviewMT} dataset.}
  \vspace{-5mm}
  \label{fig:data_processing}
\end{figure}

As shown in Figure~\ref{fig:data_processing}, the data collection process for the \texttt{ReviewMT} dataset involved gathering papers from ICLR spanning the years 2017 to 2024. Additionally, all papers published in Nature Communications in 2023 were included. For ICLR papers, we leveraged the official API~\cite{openreview_api} to systematically extract titles and abstracts. The conversion of PDF files to text was facilitated by a software tool called Marker~\cite{marker}, which ensures the text is rendered with markdown grammar, maintaining structural and formatting fidelity. For Nature Communications papers, we used the Requests library to crawl data, adhering to the Robots protocol to ensure compliance with web scraping policies. We collected papers from Nature Communications in 2023 along with their corresponding peer reviews within the official PDF files. However, some papers did not have accompanying official peer review data, and these were excluded from our dataset. All the PDF files are also converted to text with markdown grammar by Marker~\cite{marker}.

The dataset construction focused on capturing detailed and structured interactions for each turn of the peer review process. For each paper, we included several fields to support multi-turn dialogues. Specifically, the dataset includes fields for each turn of the dialogue: "Title", "Abstract", and "Main Text" provide the long context; “Summary”, “Strengths”, "Weaknesses", and "Questions" are included in the first turn for reviewers to write the initial review for a given paper; "Response" is in the second turn for authors to address each reviewer; "Final comment" and "Score" are in the third turn for reviewers to provide the final review and assign a score; and "Meta review" and "Final decision" are in the fourth turn for decision makers to make the final publication decision. All the files are stored in JSON format to ensure easy access and compatibility with various programming languages and tools.

It is important to note that the review process may vary even within the same conference across different years. For instance, some years may not provide an initial rating for the paper in the first turn, so we adapted such cases by only considering the final rating. Similarly, Nature Communications offers a “Peer Review File,” which was integrated as the main review information into our framework to ensure consistency. By meticulously collecting and organizing this data, the \texttt{ReviewMT} dataset aims to provide a comprehensive resource that captures the iterative nature of the peer review process. The resulting interactions modeled in the dataset are expected to drive more nuanced LLM applications in academic peer review, promoting constructive feedback mechanisms in scholarly publishing.

\vspace{-2mm}
\subsection{Dataset Statistics}
\vspace{-2mm}

\begin{figure}[h!]
  \centering
  \vspace{-2mm}
  \includegraphics[width=1.0\textwidth]{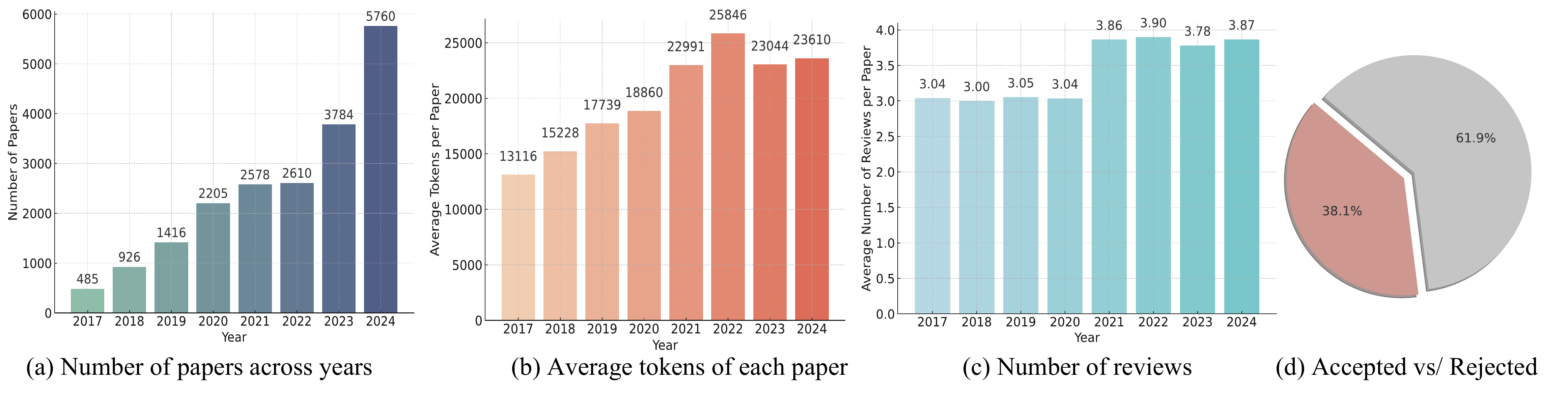}
  \vspace{-5mm}
  \caption{Statistics of the ICLR papers and reviews in the \texttt{ReviewMT-ICLR} dataset.}
  \vspace{-2mm}
  \label{fig:iclr_statistics}
\end{figure}

We provide a detailed overview of the \texttt{ReviewMT-ICLR} dataset in Figure~\ref{fig:iclr_statistics}\textcolor{red}{(a)}, which illustrates various aspects of the dataset, highlighting its significance and the challenges it addresses. The dataset showcases a remarkable increase in the number of papers submitted to ICLR, from 485 in 2017 to 5760 in 2024. This growth reflects the expanding influence and participation in the conference, underscoring the increasing importance of effective and scalable peer review processes.

Figure~\ref{fig:iclr_statistics}\textcolor{red}{(b)} depicts the average number of tokens per paper, which ranges from approximately 13,000 to 26,000 tokens. Notably, there is a slight drop in average tokens per paper in 2023 and 2024 due to the absence of rebuttal-phase replies in those years. This highlights the \textbf{long context} challenge inherent in our setting. Figure~\ref{fig:iclr_statistics}\textcolor{red}{(c)} indicates that each paper typically receives about three to four reviews. This implies that for each paper, there are at least three or four interactions between authors and reviewers. These interactions, combined with the subsequent feedback from decision makers to reviewers, form a complex \textbf{multi-turn dialogue}. Figure~\ref{fig:iclr_statistics}\textcolor{red}{(d)} presents the acceptance statistics, showing that 38.1\% of the papers in the dataset were accepted, while 61.9\% were rejected. This distribution provides a balanced mix of positive and negative samples.

\begin{figure}[h]
  \centering
  \includegraphics[width=1.0\textwidth]{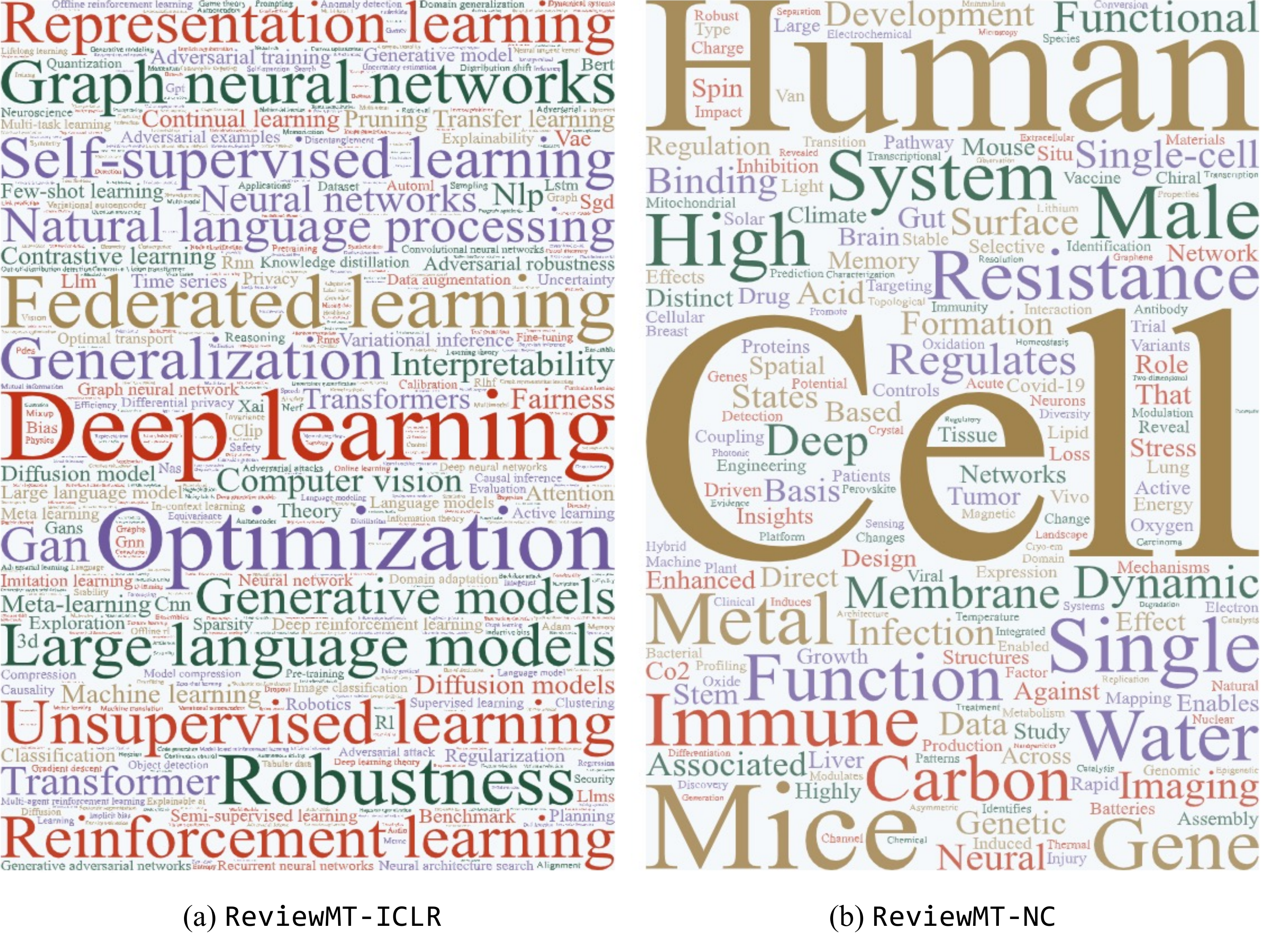}
  \caption{The word cloud of the keywords in the \texttt{ReviewMT} dataset.}
  \label{fig:iclr_word_cloud}
\end{figure}  

Figure~\ref{fig:iclr_word_cloud} presents a word cloud of keywords from the \texttt{ReviewMT} dataset, offering insights into the prevalent research themes within the dataset. The word cloud is divided into two parts, each representing keywords from different sources included in the dataset. In Figure~\ref{fig:iclr_word_cloud}\textcolor{red}{(a)}, keywords such as “deep learning”, “self-supervised learning”, and “reinforcement learning” are prominently featured. These terms highlight the focus on cutting-edge research topics in the field of machine learning that are prevalent in ICLR papers. The prominence of these keywords reflects the ongoing advancements and interests in these areas within the machine learning community. Conversely, Figure~\ref{fig:iclr_word_cloud}\textcolor{red}{(b)} illustrates a broader range of keywords derived from Nature Communications papers. These keywords cover a diverse array of scientific disciplines, including “cell”, “human”, “gene”, and “immune”. This diversity underscores the multidisciplinary nature of Nature Communications, showcasing its wide-reaching impact across various fields of scientific research. The combined keyword analysis from \texttt{ReviewMT-ICLR} and \texttt{ReviewMT-NC} demonstrates the richness and diversity of the \texttt{ReviewMT} dataset. By encompassing both specialized machine learning research and a wide spectrum of scientific disciplines, the dataset offers a comprehensive resource for training and evaluating large language models (LLMs) in the peer review process. This diversity ensures that models trained on this dataset can handle a variety of topics and review styles, enhancing their applicability and robustness in real-world academic settings.

In Table~\ref{tab:datasets}, we present detailed statistics for the \texttt{ReviewMT} dataset. The dataset encompasses a total of 26,841 papers, 92,017 reviews, and 567,207,583 tokens, calculated using the LLaMA-3 tokenizer. This extensive and diverse dataset serves as a robust resource for training and evaluating LLMs in the peer review process. The breadth and depth of the \texttt{ReviewMT} dataset ensure that it captures the complexity and nuances of real-world academic peer review, making it invaluable for this area.

\begin{table}[ht]
\small
\centering
\vspace{-5mm}
\caption{The detailed dataset statistics of the \texttt{ReviewMT} dataset.}
\setlength{\tabcolsep}{9mm}{
\begin{tabular}{lccc}
\toprule
Dataset/Year  & Papers & Reviews & Tokens \\
\midrule
ICLR 2017 & 485 & 1,474 & 6,361,501 \\
ICLR 2018 & 926 & 2,778 & 14,101,709 \\
ICLR 2019 & 1,416 & 4,322 & 25,119,319 \\
ICLR 2020 & 2,205 & 6,695 & 41,588,041 \\
ICLR 2021 & 2,578 & 9,963 & 59,272,632 \\
ICLR 2022 & 2,610 & 10,177 & 67,459,995 \\
ICLR 2023 & 3,784 & 14,307 & 87,201,458 \\
ICLR 2024 & 5,760 & 22,282 & 135,995,413 \\
\hline
ICLR & 19,764 & 71,998 & 439,100,068 \\
Nature Communications & 7,077 & 20,019 & 128,107,515 \\
\bottomrule
\end{tabular}}
\label{tab:datasets}
\end{table}

\subsection{Evaluation Metrics}

We introduce a comprehensive set of evaluation metrics tailored to the peer-review dialogue. These metrics are designed to evaluate the quality of text replies and the validity of responses.


\noindent\textbf{Validity of response} Given the long-context nature of peer-review documents, which average over 20,000 tokens per paper, LLMs may occasionally fail to provide valid responses. To address this, we use the following hit rates to evaluate the validity of the responses:
\begin{itemize}[leftmargin=5mm]
  \item \textbf{Paper hit rate}: Measures whether the LLM-generated response addresses the paper content. If the LLM fails to respond to the paper, the hit rate is 0.
  \item \textbf{Review hit rate}: Evaluates whether the LLM-generated final review includes a score. If the LLM fails to provide the score, the hit rate is 0.
  \item \textbf{Decision hit rate}: Assesses whether the LLM-generated decision includes a clear accept or reject outcome. If the LLM fails to respond with a decision, the hit rate is 0.
\end{itemize}


\noindent\textbf{Text quality evaluation} For all text replies, including the reviewers' initial reviews, the authors' responses, the reviewers' final reviews, and the decision makers' meta reviews, we employ text similarity metrics to assess the quality of the generated text. These metrics include:
\begin{itemize}[leftmargin=5mm]
  \item \textbf{BLEU-2} and \textbf{BLEU-4}~\cite{papineni2002bleu}: Measures n-gram precision by comparing the generated text to a reference text, focusing on 2-gram and 4-gram overlaps respectively.
  \item \textbf{ROUGE-1}, \textbf{ROUGE-2}, and \textbf{ROUGE-L}~\cite{lin2004rouge}: Measures f1-score of unigram, bigram, and longest common subsequence overlaps between the generated text and the reference text.
  \item \textbf{METEOR}~\cite{banerjee2005meteor}: A measure of alignment between the generated and reference texts. METEOR also incorporates stemming and synonymy, making it more sensitive to variations in wording.
\end{itemize}


\noindent{\textbf{Score evaluation}} To evaluate the accuracy of the final review scores provided by the LLMs, we use:
\begin{itemize}[leftmargin=5mm]
  \item \textbf{Mean Absolute Error (MAE)}: Measures the average absolute difference between the scores given by the LLM and the actual scores provided by human reviewers.
\end{itemize}


\noindent{\textbf{Decision evaluation}} For evaluating the decisions (accept or reject) made by the decision makers:
\begin{itemize}[leftmargin=5mm]
  \item \textbf{F1-score}: Combines precision and recall to measure the binary classification of decisions.
\end{itemize}


\section{Experiments}

We employ several open-source LLMs to evaluate performance on the proposed \texttt{ReviewMT} dataset. Specifically, we use LLaMA-3~\cite{llama3}, Qwen~\cite{bai2023qwen}, Baichuan2~\cite{yang2023baichuan}, ChatGLM3~\cite{zeng2022glm}, Gemma~\cite{team2024gemma}, DeepSeek~\cite{bi2024deepseek}, and Yuan-2~\cite{wu2023yuan}. All models are implemented using the LLaMA-factory~\cite{zheng2024llamafactory}. For \texttt{ReviewMT-ICLR}, we use papers from 2017 to 2023 as the training data and sample 100 papers from 2024 as the test set. For \texttt{ReviewMT-NC}, we sample 100 papers as the test set, with the remaining papers used for training. Both zero-shot and supervised finetuned performance are reported. Detailed implementation details and the list of sampled papers in the test set are in the supplemental materials.

\begin{table}[ht]
\vspace{-4mm}
\centering
\caption{Performance comparison of LLMs on the \texttt{ReviewMT-ICLR} dataset.}
{\renewcommand\baselinestretch{1.5}\selectfont
\resizebox{\textwidth}{!}{
\begin{tabular}{ccccccccccc}
\toprule
\multicolumn{2}{c}{Method} & Paper hit rate $\uparrow$ & Review hit rate $\uparrow$ & Decision hit rate $\uparrow$ & MAE $\downarrow$ & F1-score $\uparrow$ \\
\rowcolor[HTML]{FDF0E2} 
\cellcolor[HTML]{FDF0E2}                                   & LLaMA-3       & 100\% & 2.05\% & 9.00\% & 2.12$\pm$0.93 & 0.6154 \\
\rowcolor[HTML]{FDF0E2} 
\cellcolor[HTML]{FDF0E2}                                   & Qwen          & 89\% & 2.00\% & 58.43\% & 3.29$\pm$1.28 & 0.4068 \\
\rowcolor[HTML]{FDF0E2} 
\cellcolor[HTML]{FDF0E2}                                   & Baichuan2     & 97\% & 0.00\% & 27.84\% & /  & 0.4848 \\
\rowcolor[HTML]{FDF0E2} 
\cellcolor[HTML]{FDF0E2}                                   & Gemma         & 98\% & 1.05\% & 5.15\% & 1.25$\pm$0.43 & 0.6667 \\
\rowcolor[HTML]{FDF0E2} 
\cellcolor[HTML]{FDF0E2}                                   & DeepSeek      & 100\% & 0.51\% & 31.00\% & 4.50$\pm$1.50 & 0.6000 \\
\rowcolor[HTML]{FDF0E2} 
\cellcolor[HTML]{FDF0E2}                                   & Yuan          & 100\% & 0.00\% & 0.00\% & / & / \\
\rowcolor[HTML]{FDF0E2} 
\multirow{-7}{*}{\cellcolor[HTML]{FDF0E2} Zero-shot} & ChatGLM3      & 100\% & 19.18\% & 32.00\% & 3.36$\pm$1.92 & 0.2667 \\
\rowcolor[HTML]{E7ECE4} 
\cellcolor[HTML]{E7ECE4}                                   & LLaMA-3       & 100\% & 49.87\% & 42.00\% & 1.04$\pm$1.17 & 0.6154 \\
\rowcolor[HTML]{E7ECE4} 
\cellcolor[HTML]{E7ECE4}                                   & Qwen          & 89\% & 74.29\% & 15.73\% & 1.10$\pm$1.18 & 0.5882 \\
\rowcolor[HTML]{E7ECE4} 
\cellcolor[HTML]{E7ECE4}                                   & Baichuan2     & 99\% & 98.45\% & 14.14\% & 0.92$\pm$1.03 & 0.8000 \\
\rowcolor[HTML]{E7ECE4} 
\cellcolor[HTML]{E7ECE4}                                   & Gemma         & 98\% & 81.79\% & 48.94\% & 1.09$\pm$1.23 & 0.6522 \\
\rowcolor[HTML]{E7ECE4} 
\cellcolor[HTML]{E7ECE4}                                   & DeepSeek      & 100\% & 20.46\% & 40.00\% & 1.02$\pm$1.08 & 0.6486 \\
\rowcolor[HTML]{E7ECE4} 
\cellcolor[HTML]{E7ECE4}                                   & Yuan   & 100\% & 100.00\% & 1.00\% & 0.94$\pm$0.98 & 0.0000 \\ 
\rowcolor[HTML]{E7ECE4} 
\multirow{-7}{*}{\cellcolor[HTML]{E7ECE4}Supervised Finetune}  & ChatGLM3     & 99\% & 91.99\% & 41.41\% & 0.99$\pm$0.97 & 0.6190 \\
\bottomrule
\end{tabular}}\par}
\label{tab:iclr}
\vspace{-2mm}
\end{table}

\begin{figure}[h!]
  \centering
  \includegraphics[width=0.99\textwidth]{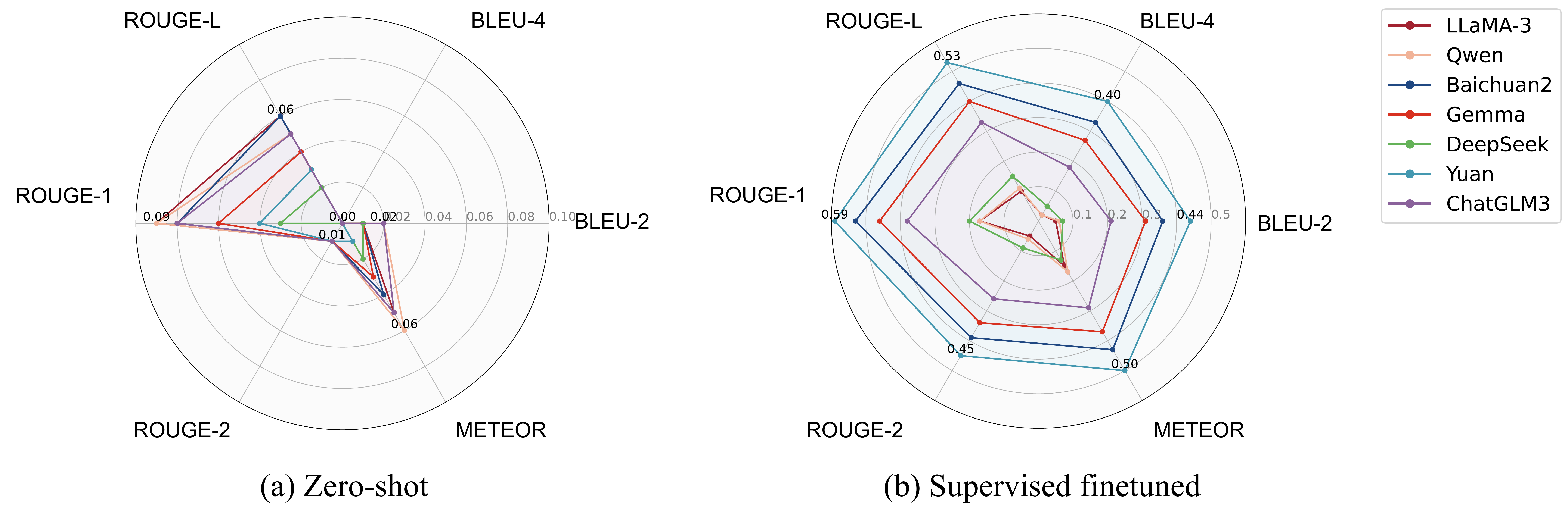}
  \vspace{-2mm}
  \caption{The radar chart of text similarity metrics for LLMs on the \texttt{ReviewMT-ICLR} dataset.}
  \vspace{-3mm}
  \label{fig:radar}
\end{figure}

As shown in Table~\ref{tab:iclr}, most LLMs demonstrate high paper hit rates, indicating their ability to generate relevant content related to the papers. However, zero-shot performance reveals lower review hit rates and decision hit rates, suggesting that LLMs struggle to provide scores and decisions. Supervised fine-tuning significantly improves performance across all metrics. For instance, the review hit rates and decision hit rates see marked improvements, reflecting the models' enhanced ability to generate comprehensive reviews and make accurate decisions after being fine-tuned on the \texttt{ReviewMT} dataset. It is noteworthy that the Yuan model, despite achieving a high review hit rate, has an extremely low decision hit rate, likely due to its strict constraints preventing it from making decisions. Figure~\ref{fig:radar} displays a radar chart comparing text similarity metrics for the evaluated LLMs. The chart illustrates that zero-shot performance is limited, whereas supervised fine-tuning yields significantly better results. High scores in text similarity metrics indicate the fine-tuned models' ability to produce text that closely matches human-generated reviews, both in terms of content and style.

\begin{table}[ht]
\small
\centering
\vspace{-4mm}
\caption{Performance comparison of LLMs on the \texttt{ReviewMT-NC} dataset.}
{\renewcommand\baselinestretch{1.5}\selectfont
\setlength{\tabcolsep}{4mm}{
\begin{tabular}{ccccccccccc}
\toprule
\multicolumn{2}{c}{Method} & Paper hit rate $\uparrow$ & Decision hit rate $\uparrow$ & F1-score $\uparrow$ \\
\rowcolor[HTML]{FDF0E2} 
\cellcolor[HTML]{FDF0E2}                                   & LLaMA-3       & 96\% & 31.25\% & 0.6364 \\
\rowcolor[HTML]{FDF0E2} 
\cellcolor[HTML]{FDF0E2}                                   & Qwen          & 37\% & 44.12\% & 0.7500 \\
\rowcolor[HTML]{FDF0E2} 
\cellcolor[HTML]{FDF0E2}                                   & Baichuan2     & 100\% & 27.00\% & 0.9811 \\
\rowcolor[HTML]{FDF0E2} 
\cellcolor[HTML]{FDF0E2}                                   & Gemma         & 99\% & 81.11\% & 0.9420  \\
\rowcolor[HTML]{FDF0E2} 
\cellcolor[HTML]{FDF0E2}                                   & DeepSeek      & 100\% & 61.00\% & 0.8269  \\
\rowcolor[HTML]{FDF0E2} 
\cellcolor[HTML]{FDF0E2}                                   & Yuan          & 100\% & 0.05\% & 1.0000 \\
\rowcolor[HTML]{FDF0E2} 
\multirow{-7}{*}{\cellcolor[HTML]{FDF0E2} Zero-shot} & ChatGLM3      & 100\% & 24.00\% & 0.8837 \\
\rowcolor[HTML]{E7ECE4} 
\cellcolor[HTML]{E7ECE4}                                   & LLaMA-3       & 100\% & 16.00\% & 0.6667 \\
\rowcolor[HTML]{E7ECE4} 
\cellcolor[HTML]{E7ECE4}                                   & Qwen          & 99\% & 21.21\% & 0.8333 \\
\rowcolor[HTML]{E7ECE4} 
\cellcolor[HTML]{E7ECE4}                                   & Baichuan2     & 100\% & 73.00\% & 0.7414 \\
\rowcolor[HTML]{E7ECE4} 
\cellcolor[HTML]{E7ECE4}                                   & Gemma         & 99\% & 89.90\% & 0.7286 \\
\rowcolor[HTML]{E7ECE4} 
\cellcolor[HTML]{E7ECE4}                                   & DeepSeek      & 100\% & 6.00\% & 0.5000 \\
\rowcolor[HTML]{E7ECE4} 
\cellcolor[HTML]{E7ECE4}                                   & Yuan   & 100\% & 18.00\% & 0.9091 \\
\rowcolor[HTML]{E7ECE4} 
\multirow{-7}{*}{\cellcolor[HTML]{E7ECE4}Supervised Finetune}  & ChatGLM3      & 100\% & 74.00\% & 0.9504 \\
\bottomrule
\end{tabular}}}
\vspace{-2mm}
\label{tab:nc}
\end{table}

\begin{figure}[h]
  \centering
  \includegraphics[width=0.99\textwidth]{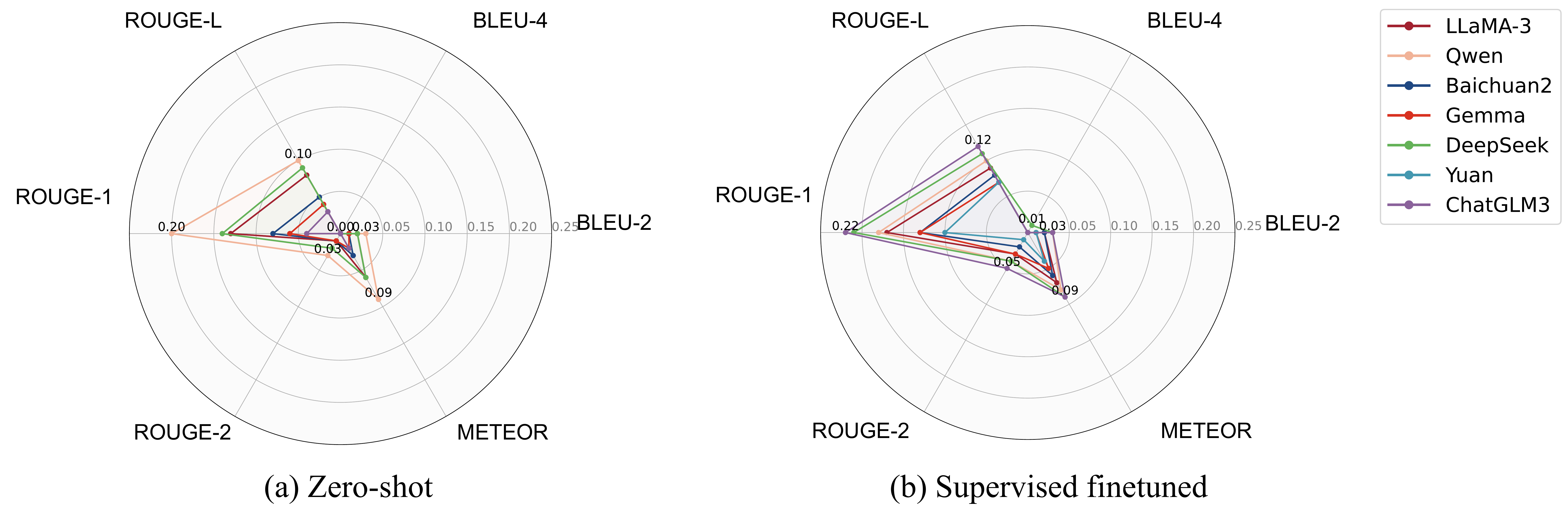}
  \caption{The radar chart of text similarity metrics for LLMs on the \texttt{ReviewMT-NC} dataset.}
  \vspace{-6mm}
  \label{fig:radar2}
\end{figure}

We present the performance results of LLMs on the \texttt{ReviewMT-NC} dataset in Table~\ref{tab:nc}. Due to the nature of the openly accessible reviews for Nature Communications, which are not as complete as those from ICLR, the evaluation focuses on a one-turn dialogue, and thus, we do not evaluate the review hit rate and MAE. Zero-shot performance suffers from low review hit rates and decision hit rates. Supervised fine-tuning improves the performance of most models. However, the extent of improvement is not as pronounced as that observed with the \texttt{ReviewMT-ICLR}. This difference could be attributed to the multi-turn dialogue nature of \texttt{ReviewMT-ICLR}, which offers a richer context for models to learn from. Figure~\ref{fig:radar2} displays a radar chart comparing text similarity metrics. The chart corroborates the results observed in Table~\ref{tab:nc}, illustrating that while supervised fine-tuning generally improves metrics, the gains are more modest compared to those seen with the \texttt{ReviewMT-ICLR}, underscoring the importance of multi-turn dialogues in the peer review process.

\vspace{-2mm}
\section{Conclusion and Limitation}
\label{sec:conclusion}
\vspace{-2mm}
In this paper, we presented the construction and evaluation of the \texttt{ReviewMT} dataset, designed for the application of LLMs in the peer review process. By reformulating peer review as a multi-turn dialogue involving distinct roles for reviewers, authors, and decision makers, we aim to capture the dynamic and iterative nature of real-world academic peer review. Our comprehensive dataset, drawn from top-tier conferences like ICLR and prestigious journals such as Nature Communications, supports this complex interaction model and provides a rich resource for fine-tuning and evaluating LLMs. Our framework includes detailed annotations for each turn of the peer review process, allowing LLMs to generate and respond to reviews. By addressing various aspects of the peer review cycle—initial reviews, author rebuttals, final reviews, and decision-making—the \texttt{ReviewMT} dataset facilitates the development of LLMs that can engage in meaningful, constructive peer review dialogues. This advancement holds promise for improving the efficiency and fairness of the peer review process.

Despite its potential, our work has certain limitations that need to be acknowledged. Firstly, no figures are included in the main text, which could limit the dataset's ability to handle visual data integral to some academic papers. Additionally, the dataset is currently limited to specific conferences and journals, primarily ICLR and Nature Communications. This scope may not fully represent the diversity of academic publishing, and future work should aim to extend the dataset to include a broader range of sources across various disciplines. The primary concern about societal impact is the potential for bias. If the training dataset includes biased reviews or decisions, the LLM might learn and replicate these biases, leading to unfair evaluations of certain groups or topics.

\bibliography{ref}
\bibliographystyle{unsrt}

\end{document}